Aashish Dhawan
Dept. of Computer Science & Engineering
JMIETI, Radaur
Yamuna Nagar, India
aashudhawan@gmail.com

Pankaj Bodani
Space Applications Center
ISRO, Bopal Campus
Ahmedabad, India
bodani.pankaj@gmail.com

Vishal Garg
Dept. of Computer Science & Engineering
JMIETI, Radaur
Yamunanagar, India
vishalgarg@jmieti.org

*Abstract*—The output of image the segmentation process is usually not very clear due to low quality features of Satellite images. The purpose of this study is to find a suitable Conditional Random Field (CRF) to achieve better clarity in a segmented image. We started with different types of CRFs and studied them as to why they are or are not suitable for our purpose. We evaluated our approach on two different datasets - Satellite imagery having low quality features and high quality Aerial photographs. During the study we experimented with various CRFs to find which CRF gives the best results on images and compared our results on these datasets to show the pitfalls and potentials of different approaches.

## I. INTRODUCTION

The process of image segmentation involves simplifying an image such as to make it meaningful and easier to analyse. Different colours are assigned to different pixels, and they are put together to form a meaningful and interpretable image. The output of segmented image forms label clusters covering the whole image, in which labels are assigned to different areas or objects in the image, as shown in Figure 1 [1]. Applications of Image Segmentation incorporate medical imaging [2], object detection [3], and object recognition.

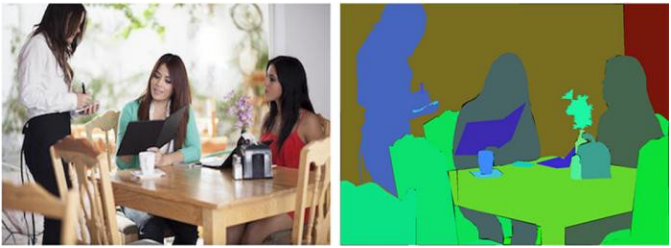

Fig. 1. Input image and its segmented image

The process of image segmentation can be a little tricky. Multiple techniques are used to segment an image. The most popular image segmentation techniques are K-Means clustering, region growing algorithms, watershed transformation, and it is even done manually sometimes. The results of these techniques may vary with the data used, the values of parameters used in the algorithm and the efficiency of the person doing the task. The accuracy is very low for satellite imagery because sometimes the images provided have low quality features due to the low resolution of the satellite providing the data. CRFs have been used previously as a Post-Processing technique for image segmentation in W-Net [4] and other Neural Network Pipelines. The purpose of this study is to find the best suitable CRF for post-processing segmented images to get highest possible accuracy.

CRFs fall into the category of Undirected Graphical Models, which follows a discriminative approach. They are often applied to the task of pattern recognition and sequence prediction. Unlike discrete classifiers, a CRF takes into account the "neighbouring" samples, example – Linear CRF are popular for Natural Language Processing (NLP) and are used for predicting arrangements of tags (labels) for a given sequence of input cases. There are multiple other types of CRF, example – Linear CRF, Grid CRF, Skip-Chain CRF, and Dense CRFs. In this paper, these CRFs are discussed along with their suitability for post-processing image segmentation.

The datasets used are taken from two different sources. The first dataset is obtained from IRS-2 sensor of LISS-3 Satellite. It contains images of Rampur, Jabalpur and a few other Indian cities. The second dataset is high resolution images provided as part of the 2D semantic labelling challenge – Potsdam, by the "International Society for Photogrammetry and Remote Sensing (ISPRS)". In both the cases, the input images as such well as labelled images are used. However, in second dataset ground truth images are also available.

CRF's have also been widely used in medical sciences and computer vision [5]. Other tasks that can be done using CRFs include shallow parsing, named identity recognition and gene finding [6].

## II. RELATED STUDY

We reviewed some work that is closely attached or relevant to ours. W-Net is a pipeline that uses CRFs as a post-processing technique to segment images. When processing any image through a neural network, surfaces become edgy on the output image. W-Net introduces a fully unsupervised approach to segment images. It uses two Convolutional Neural Networks - one for encoding and other for decoding to segment the images. Subsequently it uses fully-connected CRFs to post-process the images for edge recovery and attain better accuracy. The architecture of W-Net is shown in Figure 2 [4]. Plath et al. [7] used Support Vector Machine (SVM) implementation of the CRF for segmenting images using image features. Alam et al. [8] used a deep CRF with convolutional neural networks to segment hyperspectral images whose architecture is described in Figure 3. Other Graphical models like Markov Random Fields (MRFs) have also been used in the process of image segmentation as a post-processing technique [9].

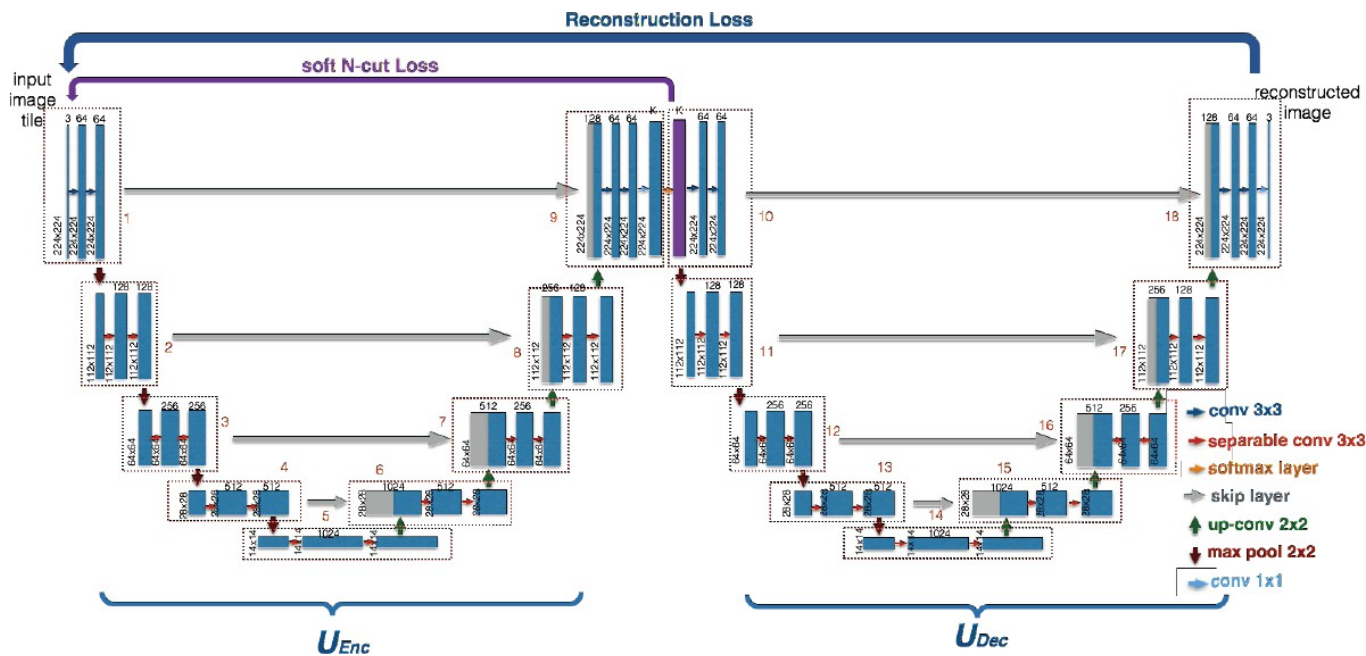

Fig. 2. Architecture of W-Net

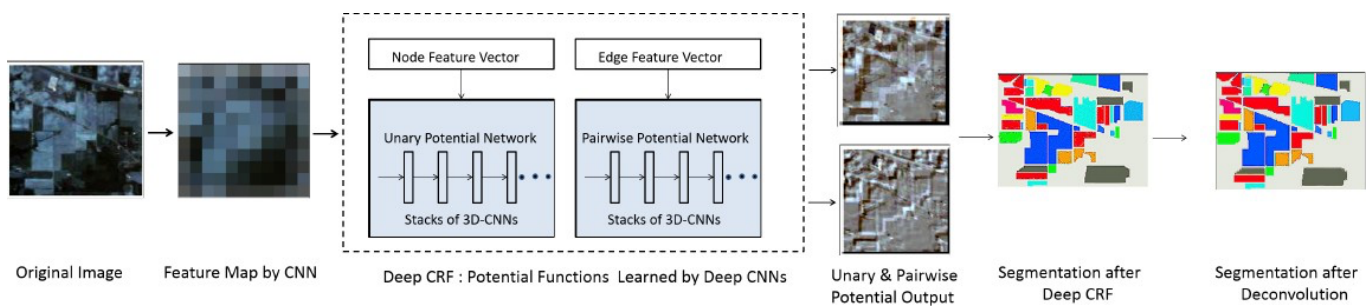

Fig. 3. Architecture of deep feature learning for hyperspectral imagery

## III. CRF MODELS

There are number of CRF models. CRFs can be utilized to show dependencies amidst input and output domains, along with the dependencies inside the domains. In generative approach, model tries to learn from the given data. Whereas in discriminative approach, model divides the data in different classes using a hyperplane as shown in Figure 4 [10].

CRFs are often used for sequence pattern recognition and structured prediction tasks. The key preferred standpoint of CRF is its capacity to incorporate subjective, non-independent features of information. It can create a direct relationship between the input and output image for better prediction. In many applications we want to predict variables that depend on each other, like the score of a student in a test or performance of a team in a particular match. Unlike other methods CRFs take "neighbouring" samples into account when making any predictions. CRFs follow conditional probability as described in Equation (1).

$$P(A|B) = \frac{P(A \cap B)}{P(B)} \quad (1)$$

P(A|B) defines the probability of event A given B has occurred. P(A∩B) is the probability of event A and B occurring together and P(B) is the probability of event B.

There are multiple types of CRFs; we started from the most basic CRF, i.e., Linear CRF, to check why/ why not it was suitable for the task of image segmentation. We tried applying all the CRF on images from both the datasets, to find the most optimal CRF to be used as a post-processing technique. In this section, we discuss all the basic types of CRFs and why/why not they were suitable for the task.

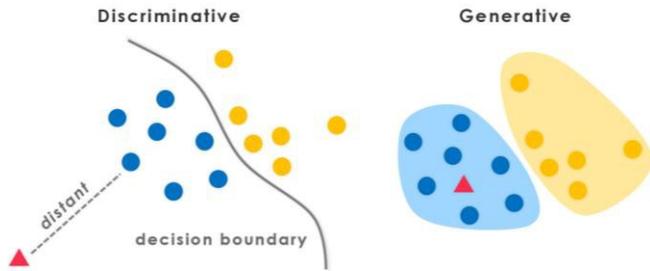

Fig. 4. Discriminative vs Generative approach

### A. Linear CRFs

As the name suggests, these are linear, and one dimensional in nature. This means that every single node is connected to its previous and next node forming a chain like structure, most commonly used for NLP tasks like part of speech tagging and shallow parsing [11] because most words are related to each other in sentences in any language.

We could not use Linear CRF in our model because of the way information is laid out in an image. The information in an image is laid in 2-dimensions and one-dimensional structure of linear CRF is unable to create relationships between the nodes/ pixels in the image. The structure of Linear CRF is shown in Figure 5 [12].

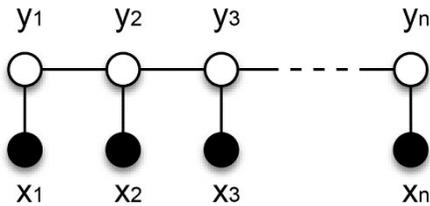

Fig. 5. Linear CRF

### B. Grid CRFs

These are two dimensional in nature, meaning every node is connected to its adjacent 4 nodes like a grid. The structure of Grid CRF is shown in Figure 6.

Grid CRFs have been used widely for pattern recognition tasks [10]. We can use these on image segmentation tasks too as shown in Figure 7 [10], but they are unable to solve complex problems because in this model, nodes are not densely connected to each other. In images, we need something more connected because of the way information is distributed in the images.

### C. Dense CRFs

Dense or Fully connected CRFs are used for more complex relationships. In this type of CRF, every node is connected to n-1 nodes, meaning all nodes in the image are connected to every other node as depicted in Figure 8.

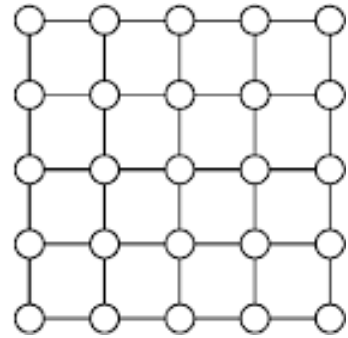

Fig. 6. Structure of Grid CRF

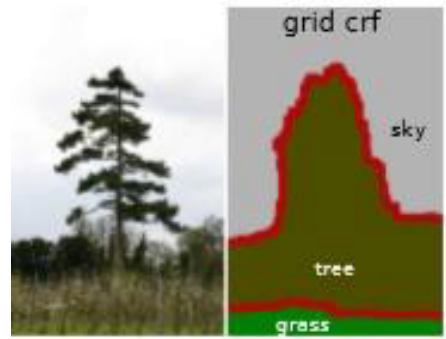

Fig. 7. Results produced using Grid CRF

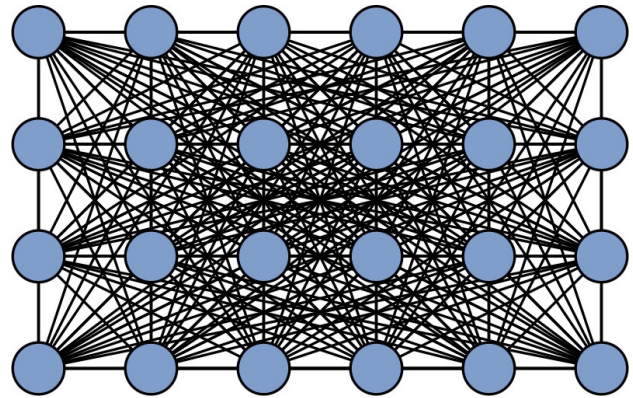

Fig. 8. Fully Connected CRF [13]

With the fully connected structure, we can interpret long range interactions that may exist in an image. This creates a perfect model for images because of the way data is laid out in an image. This is the best possible CRF to be applied on image segmentation process. The main problem with this type of CRF is that with so many relationships comes the computational complexity. To compute all these relationships, a lot of time may be required and a lot of them might not even be useful. The process on a large image such as the ones in our case can take few days to compute.

The solution was found in [14] by Krähenbühl and Koltun in their version of fully connected CRFs. It implements a highly efficient inference algorithm of fully connected CRFs. The proposed algorithm can reduce the time from 36 hours to 0.2 seconds. Using this version of fully connected CRFs is computationally less expensive and also does not make much of a difference in the quality of output. We have implemented this approach on our datasets to get the results.

IV. DATASETS

*A. City Dataset*

Space Applications Center provided the data. It contains high quality satellite images of few cities of India like Jodhpur, Rampur and others. LISS-3 sensor on the IRS-2 satellite of India captured these images. We had the original input images and the labelled images. Region Growing Algorithms were used to label this data into two classes.

*B. Potsdam Dataset*

This dataset was acquired from the "International Society for Photogrammetry and Remote Sensing (ISPRS)" for training and testing our network. The dataset relates to the city of Potsdam, which contains thirty-eight patches of 6000x6000 images taken from a sizeable true orthophoto (TOP). The images have 4 channels – red, green, blue and infrared. Each pixel in the image was categorized into one of the following classes:

- Clutter/Background (represented by the red color)
- The Car (represented by yellow color)
- Low vegetation (represented by cyan color)
- Buildings (represented by blue color)
- Tree (represented by green color)
- Impervious surfaces (represented by white color)

We used the training data from this to test our model.

V. EXPERIMENTS

We had two datasets, one with binary classification and the other with Multi-class classification. We tried creating a model that could work with any type of data i.e. Binary classified, Multi-Class classified, Hyperspectral data, and Multispectral data as well.

For the first test, we used the city dataset and applied the fully connected CRFs of Krähenbühl and Koltun [14] on data of the Rampur city. We applied the model on input and labelled image of Rampur city at Negative Probability of 70%. Figure 9 shows the dataset we had for this experiment and the initial output produced. It is a binary classified image, so it has only two specified classes - one for the yellow (urban) area and second for blue (rural).

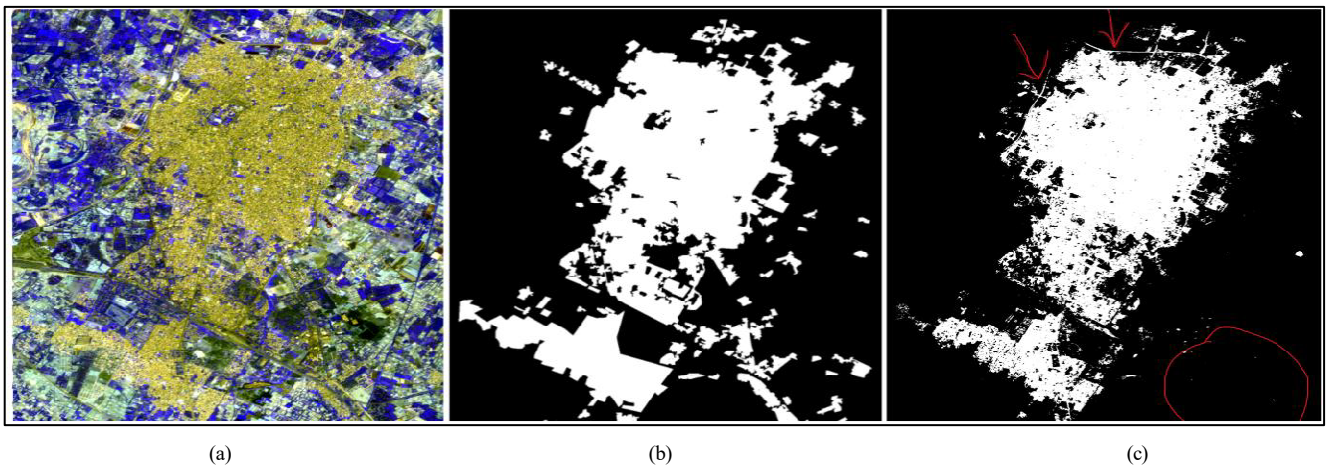

(a)  (b)  (c)

Fig. 9.  (a) Original image, (b) labelled image and (c) output produced by our model

The initial results produced were not very satisfactory. As pointed in the Figure 9 (c), using red color arrows, the under predictions were corrected and the patchy areas (over prediction) were corrected to some extent. However, it missed an area completely as marked by the red circle in the image shown above.

By altering some parameters in the algorithm like the negative probability and the potentials, we started getting better results on the data. On increasing the negative probability up to 95%, we got the results that were very good as shown in Figure 10. We could not calculate the accuracy of the output produced, as we did not have the ground truths of the image provided to us.

In the output produced, most of the errors in the prediction of the image from the initial segmentation have been corrected. Comparing it to the input images, we can see the results are satisfactory and up to the mark.

In the second dataset we had the input image, labelled images and ground truths. We tried applying the algorithm to different pairs, the first pair of the input image and labelled image and second with the labelled image and ground truth to

Post Processing of Image Segmentation using Conditional Random Fields

notice the difference in output produced and to better understand the working of fully connected CRFs.

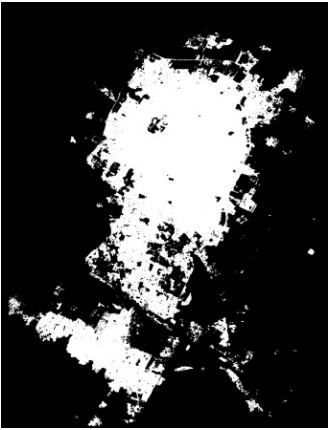

Fig. 10. Output produced at 95% negative probability

The initial results for this dataset too were not satisfactory as there was a lot of deformation in the output. The goal was to reduce the disorder in the red area and correct the wrong prediction done by the segmentation model. In the initial results, the model missed a many areas. It cleared a little bit of clutter but deformed the yellow part in the image i.e. car. The model performed well on the bus in the image. It was successful in segmenting the part which the initial model missed. However, the overall performance was not very satisfactory. The first experiment on this dataset is shown in Figure 11.

We tried correcting the results by altering the parameters in the algorithm and found some better results afterwards. The progress is shown in Figure 12. It is clearly visible that the quality of output gets better with the increase in negative probability in the algorithm. The same pattern can be seen when experimenting on the second pair of the dataset which comprises of the labelled image and the ground truth (Figure 13).

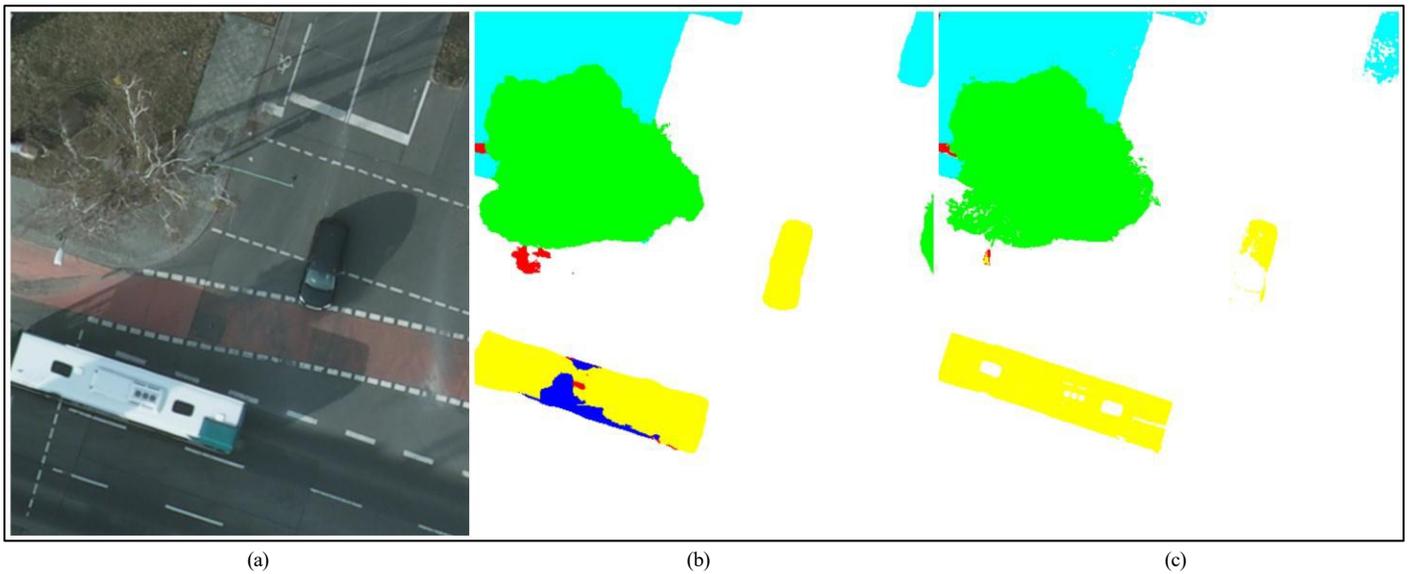

(a)                  (b)                  (c)

Fig. 11. (a) Initial image, (b) labelled image, and (c) initial result

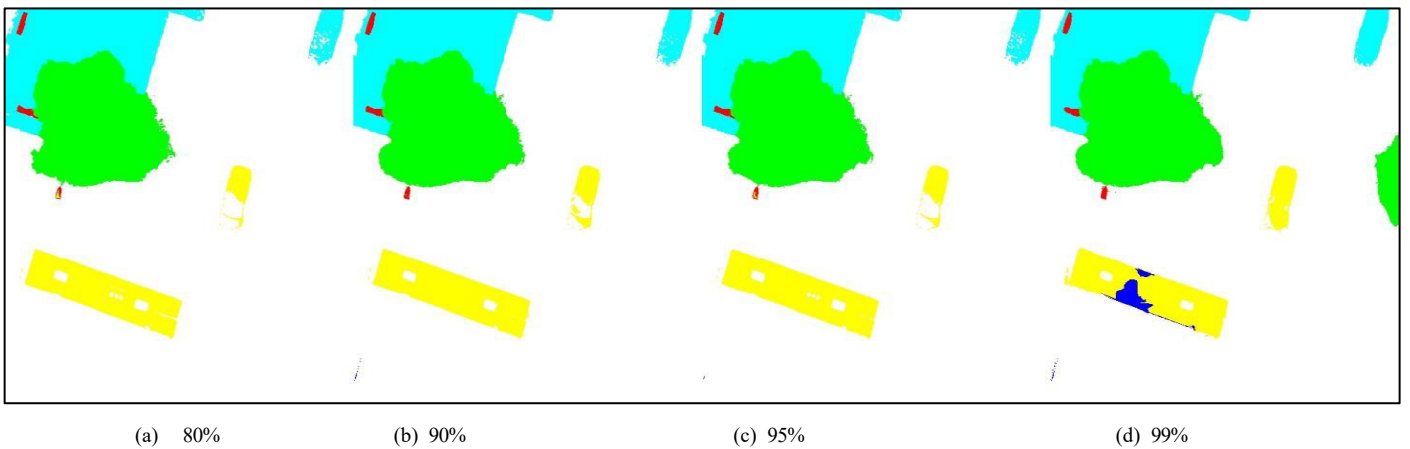

(a) 80%          (b) 90%          (c) 95%          (d) 99%

Fig. 12. Output at different negative probability levels

The results in Figure 12 show that higher the negative probability in the algorithm, better the results. The results have improved with every increase in negative probability. A similar pattern was followed in other experiments also.

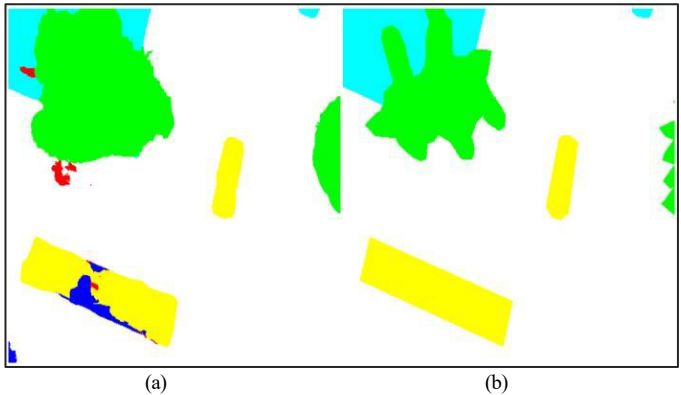

Fig. 13. (a) Labelled image, (b) ground truth

The best results produced using this data set is shown in Figure 14. The yield is very nearer to the ground truth so we can say that the model functioned admirably with the dataset.

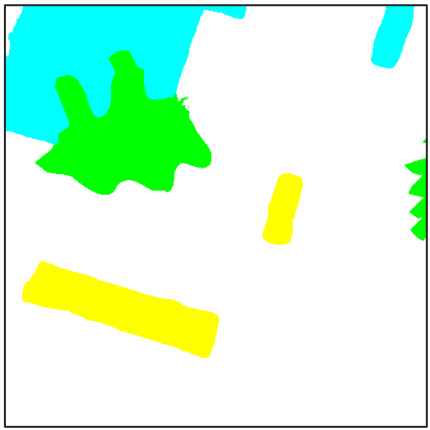

Fig. 14. Output produced using labelled image and ground truth

## VI. CONCLUSION

Although Grid and Dense CRFs can be used for processing segmented images to get higher accuracy images but they have their own pitfalls. Thus, Krähenbühl and Koltun's [14] version of fully connected CRFs is used for attaining better clarity in segmented images. The results of this algorithm improved with the increase in the amount of training data. This technique has a much greater scope in future. It can be used in developing a proper image segmentation pipeline consisting of – Pre-Processing, any unsupervised segmentation technique followed by the CRF model to get more accurate and clear segmented images.

## ACKNOWLEDGMENT

Thanks to Space Applications Center, India for giving the opportunity to work with some of the great scientists. Thanks to Dr. S. P. Vyas, Head, "ERTD (Earth-ecosystem Research and Training Division)" and Mr. Hiren Bhatt for providing a great chance to work in the TREES program and develop this model.